\documentclass{article}

 \usepackage[preprint]{neurips_2025}

\usepackage[utf8]{inputenc} 
\usepackage[T1]{fontenc}    
\usepackage{hyperref}       
\usepackage{url}            
\usepackage{booktabs}       
\usepackage{amsfonts}       
\usepackage{nicefrac}       
\usepackage{microtype}      
\usepackage{xcolor}         
\usepackage{amsmath}
\usepackage{algorithm}
\usepackage{algpseudocode} 
\usepackage{tcolorbox}
\usepackage{enumitem}

\title{Causal Distillation: Transferring Structured Explanations from Large to Compact Language Models}

\author{
  Aggrey Muhebwa\\
  Stanford University\\
  \texttt{amuhebwa@stanford.edu} \\
   \And
   Khalid K. Osman \\
    Stanford University\\
    \texttt{osmank@stanford.edu} \\
}

\begin{document}
\maketitle
\begin{abstract}
Large proprietary language models exhibit strong causal reasoning abilities that smaller open-source models struggle to replicate. We introduce a novel framework for distilling causal explanations that transfers causal reasoning skills from a powerful teacher model to a compact open-source model. The key idea is to train the smaller model to develop causal reasoning abilities by generating structured cause-and-effect explanations consistent with those of the teacher model. To evaluate the quality of the student-generated explanations, we introduce a new metric called Causal Explanation Coherence (CEC) to assess the structural and logical consistency of causal reasoning. This metric uses sentence-level semantic alignment to measure how well each part of the generated explanation corresponds to the teacher’s reference, capturing both faithfulness and coverage of the underlying causal chain. Our framework and the CEC metric provide a principled foundation for training smaller models to perform robust causal reasoning and for systematically assessing the coherence of explanations in language model outputs.
\end{abstract}
\section{Introduction}
Although causal reasoning is central to human explanation and scientific understanding, it remains underdeveloped in most language models\cite{jin2023cladder, chi2024unveiling, spirtes2000causation, chen2024causal, feder2022causal}. Large language models (LLMs) demonstrate emergent capabilities in generating structured, multi-step explanations, but these behaviors often reflect implicit pattern recognition rather than explicit modeling of causal structure \cite{zhang2023understanding, ashwani2024cause}. In smaller language models, the problem is more pronounced: the generated explanations tend to rely on shallow correlations, lacking the logical coherence and directionality that define essential cause–and–effect reasoning\cite{wang2024comprehensive, li2025survey, chi2024unveiling, cai2023knowledge}.
This limitation arises because smaller language models often lack the inductive biases necessary for causal abstraction and structured inference. As a result, they struggle to produce epistemically robust explanations that justify outcomes (faithfulness), reflect the same underlying causal mechanisms across different inputs (consistency), isolate relevant cause-and-effect components, and remain consistent under distributional shifts (generalizability).The absence of structural inductive biases for causal reasoning makes smaller language models susceptible to generating superficial explanations that are less interpretable and  sensitive to slight input variations, which undermines their reliability in tasks that necessitate systematic and explanation-driven inference.

Addressing these limitations requires a shift from purely associative prediction to models that systematically reason about the causal relationships linking evidence to claims. Current off-the-shelf LLMs rely on statistical patterns rather than causal structure, often generating plausible but logically incoherent explanations\cite{jin2023can}. For example, a model might incorrectly infer that polar bear populations are declining solely because of rising global temperatures, without considering intermediate factors such as habitat loss due to melting sea ice. Thus, there is a need for models that extend beyond surface-level correlations and are capable of reasoning about how evidence supports or refutes specific claims.

In this work, we introduce a novel approach to causal knowledge distillation: a supervised fine-tuning (SFT) framework that utilizes causal knowledge distillation to embed causal reasoning capabilities into smaller LLMs\cite{hinton2015distilling, fukuda2017efficient, wu2021causal}. We use a proprietary LLM (GPT-4) as the teacher model to generate structured cause-and-effect explanations that connect claims with their supporting evidence. These explanations are then used to fine-tune smaller open-source student models (e.g., TinyLlama, Gemma-2B) by training them to mirror the teacher’s causal reasoning.

To transfer causal reasoning capabilities from a larger language model to a smaller one, we adopt a distillation-based framework in which the student model is trained to reproduce the structured explanations generated by a more capable teacher model\cite{hinton2015distilling}. Given an input consisting of a claim and supporting evidence, the teacher model generates a causal explanation that serves as a supervision signal for the student model. To ensure that the student model generates explanations consistent with those of the teacher model, the student is fine-tuned using a divergence-based objective. This objective, such as Kullback–Leibler divergence, is computed over the training distribution and guides the student to replicate the teacher’s reasoning process. As a result, the student model learns to produce explanations that are causally structured and aligned with the underlying logic of the teacher model.

While our causal distillation framework enables smaller models to generate more meaningful explanations, evaluating the quality of these explanations presents an additional challenge. Existing metrics for evaluating generated text (e.g., BERTScore, BLEU, and ROUGE) primarily focus on surface-level similarity or general semantic alignment\cite{yao2023human, hase2020leakage}. In doing so, they fail to adequately capture whether an explanation provides a logically coherent chain of causal reasoning grounded in evidence. As a result, these metrics provide superficial assessements based on lexical or semantic overlap, but fail to differentiate between explanations that appear coherent and those that capture meaningful causal structure. To address this challenge, we introduce Causal Explanation Coherence (CEC), a novel evaluation metric designed to measure how effectively a model’s explanation captures the underlying cause-and-effect relationships that connect the evidence to the claim. Intuitively, an explanation with high CEC is one that is internally consistent, logically follows a $ \text{cause} \rightarrow \text{effect} $ progression, and accurately links evidence to the claim's outcome. CEC is grounded in the cognitive psychology theory of explanation coherence, which refers to the extent to which causal elements form a logically connected and mutually supportive narrative\cite{thagard1989explanatory}. Thus,  we implement CEC by comparing the semantic structure of a model’s explanation with that of a trusted reference which in this case, comes from a strong proprietary (e.g., GPT-4) teacher model. CEC evaluates how closely each part of the generated explanation aligns with causally relevant segments of the reference and vice versa, rewarding explanations that convey the same underlying causal relationships even when expressed with different wording.
Our contributions are summarized as follows:
\begin{itemize}
    \item \textbf{Causal Distillation for Explanations}: We propose a supervised fine-tuning framework that transfers causal explanation capabilities from a large proprietary teacher model to smaller open-source student models, enabling the students to generate coherent, evidence-based explanations for claim–evidence pairs.
    \item \textbf{Causal Coherence Metric}: We introduce the Causal Explanation Coherence (CEC) metric to evaluate how well an explanation captures the underlying cause-and-effect relationships supported by the evidence at hand.
    \item We demonstrate that CEC correlates more consistently with human-judged explanation quality than existing metrics, establishing it as a principled and interpretable measure for explanatory NLP tasks.
\end{itemize}
In the sections that follow, we position our work within the broader context of related research (Section\ref{sec:related_work}), outline our methodology for causal knowledge distillation and the CEC metric (Section\ref{sec:methodology}), and conclude with a discussion of the approach’s implications, limitations, and directions for future work (Section\ref{sec:discussion}).

\section{Related Work}
\label{sec:related_work}
Our approach builds on and intersects with several areas of existing research, including causal reasoning in language models and evaluation metrics for generated text in NLP.
\textbf{Causal Reasoning in LLMs}: While LLMs perform competitively on standard reasoning benchmarks, their outputs often depend on in-distribution statistical correlations rather than strong, generalizable reasoning strategies.
This limitation is particularly evident in compact language models. When evaluated under distributional shifts or unfamiliar reasoning scenarios, these models show significant performance degradation as explanations often fail to provide coherent, stepwise causal inference,  relying instead on the spurious correlations learned during pretraining. While prompting strategies such as chain-of-thought can improve performance on simple causal questions, they do not consistently allow for deeper causal abstraction or generalization.
Recent studies show that even the most advanced LLMs struggle with complex or counterfactual causal queries\cite{chen2025counterbench}. For example, a model may generate a fluent explanation that attributes causality to an irrelevant or logically flowed event, which reveals a failure to internalize causal structure. This limitation stems from the fact that transformer-based LLMs are trained using next-word prediction, where the model learns to predict the most likely subsequent token in a sequence based on prior context. While this method is effective for surface-level fluency, it does not guarantee an understanding of real-world cause-and-effect relationships. Out-of-the-box LLMs may therefore produce explanations that appear coherent but fail to establish a logical connection between evidence and outcome.
While recent research has explored training  bigger language models for causal reasoning\cite{openai2025o3mini, anthropic2025claude37, google2025geminiflash, abdin2025phi, guo2025deepseek}, most approaches have focused on general logical consistency or commonsense inference rather than explicitly modeling structured, evidence-based cause-and-effect relationships. Furthermore, the evaluation of such reasoning remains underdeveloped, with few metrics available to assess the coherence of causal explanations.

\textbf{Knowledge Transfer via Explanation Distillation}: Knowledge distillation (KD) is a model compression technique in which a larger and more complex model (teacher) transfers knowledge to a smaller (student) model by training the student to replicate the teacher’s outputs (behavior)\cite{hinton2015distilling}. Traditional knowledge distillation compresses knowledge by training a student model to match the teacher’s output distributions (soft labels). Recent extensions incorporate explanation distillation, where the student model is also guided by the teacher’s intermediate reasoning steps or natural language explanations rather than final predictions alone \cite{hsieh2023distilling, magister2022teaching}. This approach aims to transfer both the task-specific knowledge and the reasoning processes underlying the teacher model's outputs. However, most distillation efforts continue to focus on improving task accuracy, with limited emphasis on the quality, structure, or faithfulness of explanations. One major challenge with using these large and complex models such as GPT-4 as teachers is the limited access to their internal knowledge representations. In such cases, the generated explanations serve as the only available supervision signals, acting as indirect proxies for the teacher’s underlying reasoning. This presents both an opportunity and a challenge. While high-quality, model-generated explanations can bootstrap training when human-annotated data is unavailable, they may also encode biases or inconsistencies in the teacher’s reasoning process.

\textbf{Explanation Evaluation Metrics}: Evaluating the quality of generated explanations presents a significant challenge in NLP. Evaluation metrics based on n-gram overlaps, such as BLEU or ROUGE, often fail to accurately represent an explanation’s true quality by penalizing a valid explanation simply for using different wording than the reference. For instance, an explanation could be factually accurate and well-reasoned but still receive a low ROUGE score due to minimal lexical overlap with the reference explanation. Recent research has shifted towards embedding-based metrics (e.g., BERTScore, BLEURT), which assess semantic similarity and generally show a stronger correlation with human judgments of explanation relevance and correctness. Although these metrics are more forgiving of wording variations and better at capturing the essence of an explanation, they still do not specifically address causal relationships. In particular, a high BERTScore does not guarantee that the explanation accurately connects causes (evidence) to effects (claim outcomes); it merely indicates overall semantic similarity. While these metrics are useful for evaluating generated text, they do not provide a unified, interpretable measure of how well an explanation captures causal relationships, which is an important requirement in misinformation detection.

\section{Methodology}
\label{sec:methodology}
\subsection{Causal Distillation Framework}
\textbf{Problem Setup}: We consider inputs $X$, which consist of a claim paired with supporting evidence (e.g., a claim about climate science and a relevant passage from Wikipedia). A pretrained LLM without specialized training could typically identify whether the claim is true or false by recognizing associative patterns, but would not articulate the causal reasoning. GPT-4 serves as an implicitly causal-aware teacher model $M_{\text{teacher}}$, and then for each training instance $X$, we obtain an explanation $M_{\text{teacher}}(X)$ that explains why the evidence supports or contradicts the stated claim. These teacher-generated explanations provide consistent causal rationales that serve as quality supervision for training student models in the absence of human-annotated explanations. We then fine-tune a smaller open-source model $M_{\text{student}}(X; \theta)$ (with parameters $\theta$) using this data so that it learns to generate similar explanations. The training objective is to minimize the difference between the student model’s explanation and the teacher’s explanation across all training instances. Formally, we minimize a causal explanation divergence objective:

\begin{equation}
\theta^* = \arg\min_{\theta} \mathbb{E}_{X \sim p(X)} \left[ D\left( M_{\text{teacher}}(X), M_{\text{student}}(X, \theta) \right) \right]
\label{eq:teacher_student}
\end{equation}
where $D(\cdot,\cdot)$ measures the dissimilarity between two explanations. In practice, $D$ can be implemented as a token-level loss (e.g., cross-entropy if treating the teacher’s explanation as the target text) or as a semantic distance (e.g., minimizing the cosine distance between the teacher and student explanation embeddings). Optimizing Equation \ref{eq:teacher_student} allows the student model to learn to replicate the causal reasoning patterns demonstrated by the teacher model. To achieve this, we fine-tune the student model using standard SFT techniques on a benchmark dataset of claim–evidence pairs, each accompanied by an explanation created by GPT-4. These explanations guide the student model in determining whether a claim is supported or contradicted, using coherent, evidence-based causal reasoning. In our implementation, we fine-tune three student LLMs (TinyLlama-1.1B, Phi-2 1.3B, and Gemma-2B) on the Climate-FEVER dataset enriched with GPT-4 explanations, using consistent data points, data splits, and hyperparameters. This process bridges associative and causal reasoning: the student models retain their original language understanding and classification capabilities while gaining the added ability to explain decisions in causal terms. 

\subsection{The Causal Explanation Coherence Metric}
\label{subsec:cec}
Standard NLP metrics (e.g., BERTScore) fail to capture whether a generated explanation preserves the underlying causal reasoning that links the evidence to the claim. To address this, we introduce  thr CEC metric, which quantifies the semantic alignment between a generated explanation and a causally grounded reference. CEC evaluates whether both explanations express the same cause-and-effect relationships while allowing for variations in phrasing and structure. We define the CEC score for a single instance as follows:

Let $E_{\text{gen}} = {g_1, g_2, \dots, g_n}$ be the set of sentences in the generated explanation, and $E_{\text{ref}} = {a_1, a_2, \dots, a_m}$ represent the sentences in the reference explanation. We first encode each sentence into a dense vector representation (using a sentence embedding model) to capture its semantic content. Denote $g_i \in \mathbb{R}^d$ as the embedding of the $i$-th generated sentence and $a_j \in \mathbb{R}^d$ as that of the $j$-th reference sentence. We then compute a bidirectional semantic similarity: for each generated sentence, we identify the reference sentence that is most similar to it (the one with the highest cosine similarity), and do the same for each reference sentence. The CEC score is then defined as the average of these maximum similarities from both sides, providing a symmetric measure of alignment:
\begin{equation}
\text{CEC}_{\text{sym}}(E_{\text{gen}}, E_{\text{ref}}) = 
\frac{1}{2} \left( 
\frac{1}{n} \sum_{i=1}^{n} \max_{1 \leq j \leq m} 
\frac{\mathbf{g}_i^\top \mathbf{a}_j}{\|\mathbf{g}_i\| \|\mathbf{a}_j\|} 
+ 
\frac{1}{m} \sum_{j=1}^{m} \max_{1 \leq i \leq n} 
\frac{\mathbf{a}_j^\top \mathbf{g}_i}{\|\mathbf{a}_j\| \|\mathbf{g}_i\|} 
\right)
\end{equation}

This formulation yields a score between 0 and 1 (when using cosine similarities) that indicates how well each sentence in the generated explanation aligns with a corresponding sentence in the reference, and vice versa. The first term of the equation (forward direction) measures coverage by evaluating whether each part of the generated explanation aligns with a relevant section of the reference. This ensures that the generated explanation includes all critical causal points from the ground truth. The second term (backward direction) assesses faithfulness and completeness by determining whether each essential causal element in the reference explanation is adequately represented in the generated explanation. By averaging the forward and backward directions, the symmetric CEC score captures both types of misalignment: it penalizes missing causal elements that should have been included in the generated explanation and additional content that lacks justification in the reference explanation. Importantly, this metric is order-invariant and robust to paraphrasing: the sentences do not need to follow the same order or use identical wording, as long as the critical causal content is retained. For example, if the reference explanation includes two distinct causal statements, A and B, and the explanation generated by the model includes both (regardless of whether they are phrased differently or presented in a different sequence), the CEC score will remain high. In contrast, if the model omits B or introduces a causal link that is not supported by the reference, the score will decrease to reflect this misalignment.

\textbf{Interpretation}: A high CEC score indicates that the model’s explanation captures the core cause-and-effect relationships presented in the reference, suggesting alignment with the underlying causal structure that connects evidence to the claim. In contrast, a low CEC score suggests that the explanation either overlooks essential causal links or introduces reasoning that is unsupported or irrelevant. In comparison to one-directional or surface-level lexical metrics, CEC offers a more reliable measure of explanation quality by capturing bidirectional causal alignment between generated and reference explanations. CEC directly evaluates whether the causal elements in an explanation are logically consistent and mutually supportive, addressing a core requirement of structured reasoning. This distinction is important in tasks that require more than surface fluency, where explanations must faithfully capture how evidence supports or contradicts a claim through coherent cause-and-effect relationships.

\begin{algorithm}[ht]
\caption{Causal Explanation Coherence (CEC) for a single claim-evidence Instance}
\label{alg:symmetric_cec}
\begin{algorithmic}[1]
\Require Generated explanation sentences $E_{\text{gen}}=\{g_1, \dots, g_n\}$,
Reference explanation sentences $E_{\text{ref}}=\{a_1, \dots, a_m\}$, embedding function $\text{Embed}(\cdot)$
\Ensure Symmetric CEC score for the given instance

\State $\mathbf{G} \gets [\text{Embed}(g_1), \dots, \text{Embed}(g_n)]$ 
\State $\mathbf{A} \gets [\text{Embed}(a_1), \dots, \text{Embed}(a_m)]$ 

\State Initialize lists $S_{\text{forward}} \gets []$, $S_{\text{backward}} \gets []$

\For{$i \gets 1$ \textbf{to} $n$}
    \State Compute similarities: $\text{sim}_i \gets [\cos(\mathbf{g}_i, \mathbf{a}_j) \ \text{for} \ j = 1 \dots m]$
    \State $S_{\text{forward}}.\text{append}(\max(\text{sim}_i))$
\EndFor

\For{$j \gets 1$ \textbf{to} $m$}
    \State Compute similarities: $\text{sim}_j \gets [\cos(\mathbf{a}_j, \mathbf{g}_i) \ \text{for} \ i = 1 \dots n]$
    \State $S_{\text{backward}}.\text{append}(\max(\text{sim}_j))$
\EndFor

\State $\text{CEC}_{\text{forward}} \gets \frac{1}{n} \sum_{i=1}^{n} S_{\text{forward}}[i]$
\State $\text{CEC}_{\text{backward}} \gets \frac{1}{m} \sum_{j=1}^{m} S_{\text{backward}}[j]$
\State $\text{CEC}_{\text{sym}} \gets \frac{1}{2} (\text{CEC}_{\text{forward}} + \text{CEC}_{\text{backward}})$

\State \Return $\text{CEC}_{\text{sym}}$
\end{algorithmic}
\end{algorithm}

\section{Experiments}
\subsection{Experimental Setup}
\label{subsec:exp_setup}
\textbf{Dataset}: We evaluate our framework using the Climate-FEVER dataset, a data benchmark designed for misinformation detection in climate science, which includes 1,535 real-world climate-related claims\cite{diggelmann2020climate}. Each claim is paired with up to five evidence sentences sourced from Wikipedia and is categorized as "supported", "refuted", or "not enough info". However, Climate-FEVER does not provide human-written explanations for why a claim is true or false. We address this limitation by using GPT-4 as a teacher to generate causal explanations for each claim-evidence pair. These teacher-generated explanations describe how the evidence either supports or contradicts the claim, highlighting the causal relationship between the evidence and the claim’s accuracy. This process creates a synthetic training set of claim-evidence-explanation instances. Each instance is structured as follows: [Claim]; [Evidence]; [Label]; along with an open-ended prompt for the explanation. The GPT-4 explanation serves as the reference “ground truth” for that instance.

\textbf{Teacher and Student Models}: We use GPT-4 as the teacher model because of its strong implicit reasoning capabilities. For the student models, we selected three open-source LLMs with different sizes and architectures: TinyLlama (1.1B parameters), Phi-2 (approximately 2B), and Gemma-2B (2.2B)\cite{gemma_2024, zhang2024tinyllama, javaheripi2023phi}. Each of these is significantly smaller than GPT-4 and falls into the category of compact models (under 3B parameters) that we aim to equip with causal reasoning abilities. These models were selected to evaluate whether diverse, smaller architectures can learn to generate coherent causal explanations after fine-tuning. Each student model was initialized from its pre-trained weights and then fine-tuned on the climate claim-evidence data, using GPT-4 explanations as targets.

\textbf{Training Details}: We conducted supervised fine-tuning to generate causal explanations using next-token prediction (teacher-forcing) over concatenated inputs \texttt{[Claim; Evidence; Label]}. The objective reflects causal knowledge distillation, training student models to mimic teacher explanations. To reduce memory overhead and enable efficient training, we applied 4-bit quantization and low-rank adaptation (LoRA)\cite{hu2022lora}. All models were trained for 10 epochs on Google Cloud Platform using two NVIDIA GPUs, with a learning rate of $2e^{-5}$, batch size of 8, and the maximum sequence length supported by each model.

\subsection{Main Results: Explanation Quality and Causal Coherence}

\begin{table}[h!]
\centering
\caption{Comparison of LLMs across Evaluation Metrics}
\label{tab:model_comparison}
\small
\begin{tabular}{lccccccc}
\toprule
Model & BLEU & ROUGE-1 & ROUGE-2 & ROUGE-L & METEOR & BERTScore & CEC \\
\midrule
Phi-2 & 0.247 & 0.556 & 0.312 & 0.385 & 0.492 & 0.722 & 0.910 \\
TinyLlama-1.1B & 0.200 & 0.518 & 0.253 & 0.338 & 0.436 & 0.683 & 0.895 \\
Gemma-2-2B & 0.104 & 0.356 & 0.130 & 0.221 & 0.286 & 0.622 & 0.860 \\
\bottomrule
\end{tabular}
\end{table}

 Table \ref{tab:model_comparison} summarizes the results of the generated explanations for each model, evaluated against the GPT-4 teacher explanations (ground truth). For each model, we report BLEU, ROUGE-1/2/L, METEOR, BERTScore, and our CEC metric.
 
Our experiments demonstrate that the proposed causal distillation approach improves the ability of the  smaller language models to generate coherent and logically sound explanations for causal relationships. The distilled models consistently achieve high causal coherence scores (CEC \( \geq \) 0.86), with Phi-2 showing the strongest performance at 0.910. These scores confirm that causal distillation effectively transfers complex causal reasoning capabilities from the larger teacher model to the compact student models.

A deeper analysis of evaluation metrics shows that conventional lexical overlap metrics like BLEU and ROUGE consistently undervalue the quality of generated explanations from smaller models. For example, Gemma-2B achieves a low BLEU score (0.104) despite a high CEC score (0.860), ideally indicating strong causal alignment. This pattern is evident across other small models, suggesting a systematic divergence between lexical and coherence-based evaluations.

While embedding-based metrics like BERTScore capture more semantic nuance than traditional lexical metrics, our results indicate that they may still overestimate explanation quality in the absence of causal coherence. For example, Gemma-2B achieved a moderate BERTScore of 0.622, which is higher than its BLEU score of 0.104 and ROUGE-L score of 0.221, suggesting a degree of semantic alignment. However, the same model achieved a high CEC score of 0.860, demonstrating strong causal coherence despite relatively low performance on surface-form metrics. Similarly, TinyLlama-1.1B attained a BERTScore of 0.683 and a CEC of 0.895, while Phi-2 scored 0.722 and 0.910, respectively. These patterns suggest that BERTScore, although better than BLEU and ROUGE, still struggles to fully distinguish between simply topically similar explanations and those that show accurate cause-and-effect reasoning. In contrast, CEC differentiates causal fidelity from semantic proximity more effectively, offering a complementary signal for evaluating explanations.

\subsection{Statistical Significance of Metric Comparisons}
\label{subsec:significance_tests}
Since all metrics were calculated using the same set of explanations, we randomly selected 100 pairs of explanations and performed paired-sample t-tests to determine whether the CEC metric produces scores that significantly differ from those of current metrics. We compared the CEC metric with BERTScore, as both metrics use contextualized embeddings to evaluate semantic similarity. However, CEC operates at the sentence level to assess causal alignment, while BERTScore focuses on token-level correspondence for overall text quality. The results show statistically significant differences between CEC and the comparison metric ($p < .001$). Specifically, the average CEC score ($M = 0.908, SD = 0.017$) was significantly higher than the average BERTScore ($M = 0.720, SD = 0.047$), $t(99) = 100.34$, $p < .001$, with an exceptionally large effect size ($d = 5.18$). While this effect size is unusually high, it reflects the consistency with which CEC scores exceed BERTScore across individual explanation pairs. Additionally, we conducted a non-parametric Wilcoxon signed-rank test, which yielded a test statistic of $W = 0.0$ and $p < .001$. 
These statistical results confirm that our proposed metric consistently assigns higher and more distinctive scores to causally coherent explanations, demonstrating greater sensitivity to causal reasoning quality than conventional semantic or lexical metrics.

\subsection{Qualitative Analysis: Example Explanations}

\begin{figure}[h]
  \centering
  \includegraphics[width=1.0\textwidth]{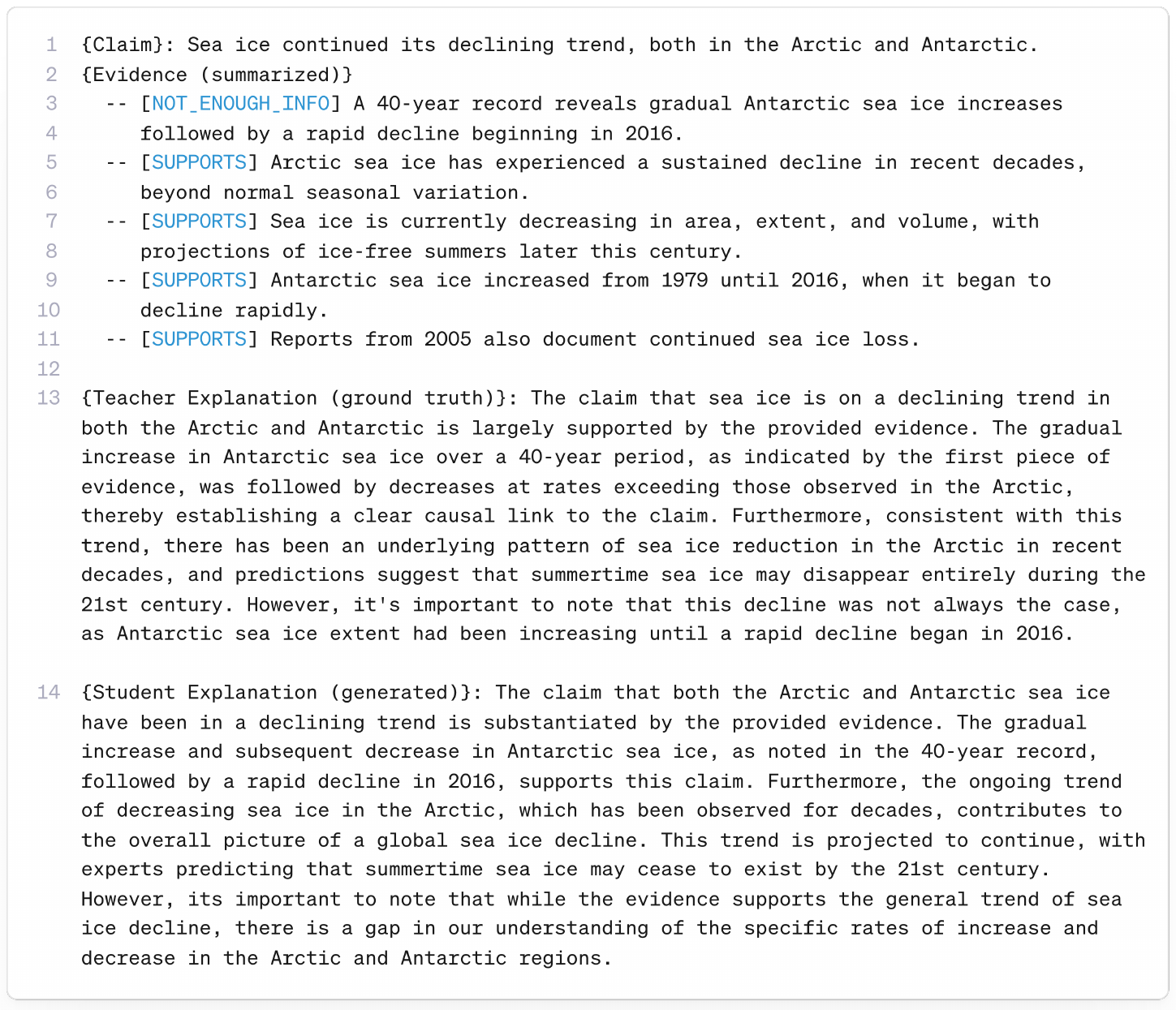}
  \caption{The student model’s explanation preserves the causal structure of the teacher’s output and maintains coherence across multiple evidence sources. However, it simplifies regional contrasts and omits some of the temporal details.}
  \label{fig:example_of_text}
\end{figure}

To illustrate the structure and quality of the generated explanations, we present a selected example that compares the outputs from the student model and teacher model (Figure~\ref{fig:example_of_text}). This example involves a complex temporal claim supported by multiple, partially overlapping sources of evidence that span different geographic regions and time periods, and demonstrates the fine-tuned student model's ability to capture key causal elements while maintaining coherence.

\vspace{0.1em}
In a comparative analysis, both teacher and student models generated explanations that incorporated evidence from multiple regions and time periods. The teacher model explicitly contrasted Antarctic behavior before and after 2016, highlighting regional differences in sea ice decline rates. The student model identified the key temporal breakpoint (2016) and acknowledged the shift in Antarctic trends, demonstrating its ability to replicate the logical structure of the teacher's explanation.

\subsection{Failure Modes and Limitations}
An inspection of the model outputs across multiple examples reveals several recurring failure patterns in the student models distilled from the teacher model. A primary limitation is the generation of unsupported or incomplete causal chains. While student models often identify the correct pieces of evidence, they frequently overlook intermediate reasoning steps or introduce irrelevant details. For example, a student model might correctly recognize elevated atmospheric $CO_2$  levels but fail to explain the underlying mechanisms linking these levels to the rise in global temperature. Another common error involves misinterpreting causal directionality, where the student model reverses the true cause-and-effect relationship. In such cases, the model may incorrectly suggest that rising global temperatures lead to increased $CO_2$  emissions, which contradicts the established causal order. These patterns indicate that while causal distillation improves surface-level coherence, deeper causal reasoning remains a significant challenge for compact models.
\section{Discussion}
\label{sec:discussion}
Our framework presents a novel approach to distilling causal explanations, where a compact student model is trained to emulate the structured, multi-step reasoning exhibited by a larger teacher model (e.g., GPT-4). Unlike traditional knowledge distillation that largely focuses on aligning output probabilities or label predictions, our method focuses on transferring explanatory structure, thereby capturing not just what the model predicts but also why. thus, the student learns to encode coherent chains of causal inference, acquiring interpretable and epistemically grounded representations of cause-and-effect relationships. This shift from predictive mimicry to reasoning transfer represents a conceptual advancement in causal language modeling, enabling more transparent, efficient, and accessible models that retain the explanatory depth of larger proprietary or closed-source models.

To evaluate this capability, we introduced the CEC metric, a novel measure designed to assess whether a model-generated explanation semantically aligns with the causal relationships in a trusted reference. Instead of focusing on surface-level or token-level similarity, CEC employs bidirectional sentence-level semantic alignment to approximate the preservation of causal structure between the generated and reference explanations. This provides a more targeted and interpretable signal for assessing explanation quality, aligning the evaluation process more closely with the objective of reasoning transfer.

That said, the student models reveal consistent limitations in how they encode and express causal reasoning. In particular, they tend to simplify fine-grained causal distinctions, merging region-specific or temporally localized insights into broader narratives. While such generalization reflects the limitations of restricted model capacity, it also establishes a boundary: causal distillation can preserve core reasoning patterns but may reduce explanatory granularity. Nonetheless, the approach remains domain-agnostic. Although our experiments focus on climate misinformation, the framework can generalize naturally to other tasks such as legal justification, scientific question answering, and clinical decision support, all which require structured explanatory behavior.

\textbf{Limitations}: The distillation process fundamentally depends on the quality of the teacher’s explanations. This means that any inaccuracies, omissions, or biases in the causal reasoning produced by the teacher model could be faithfully replicated by the student model. As a result, the student model does not discover causal structures independently but rather imitates a fixed explanatory paradigm, which raises the risk of overfitting to the teacher’s reasoning style or domain-specific patterns. While the student models generalize well within the data distribution observed during training, they often struggle with out-of-distribution examples that require novel causal chains or multi-hop reasoning steps beyond what the distillation process exposed them to. 

Another limitation lies in our evaluation framework. The CEC metric is designed to assess the internal logical structure and semantic alignment between a student model’s explanation and a trusted reference. However, CEC does not verify factual correctness. As a result, a model may generate a causally coherent yet factually incorrect explanation and still receive a high CEC score, potentially obscuring critical errors. Additionally, CEC has interpretability limitations: it reduces explanation quality to a single scalar value without identifying which causal components influenced high or low scores. This limits its utility for diagnostic analysis and model improvement. Therefore, while our proposed approach provides a structured and principled approach to modeling causal explanations, it is not a comprehensive solution. Our modeling approach and the CEC evaluation metric represent a foundational step toward achieving causal interpretability in language models. The proposed CEC metric evaluates a model’s internal coherence and causal alignment, supplementing traditional reliability methods such as human evaluation and factual assessment. Unlike metrics that focus solely on lexical overlap or factual accuracy, CEC evaluates the structure of causal reasoning in model-generated explanations, providing a novel and complementary perspective on explanation quality. This enables a more precise evaluation of explanation quality across various tasks and domains, ultimately advancing the development of more causally interpretable language models.

\textbf{Directions for future work}: While the current training process promotes overall causal coherence, future iterations could incorporate targeted objectives, such as auxiliary loss terms, contrastive supervision, or attention regularization, to retain critical explanatory details that are often lost during compression. These improvements could enable student models to better approximate the depth and specificity of reasoning generated by teachers. On the evaluation side, there is significant potential to expand and refine the CEC metric. One approach is to augment CEC with factuality checks, ensuring that high coherence does not mask explanatory hallucinations or incorrect premises. Alternatively, models designed to identify weaknesses in causal logic could provide a more comprehensive assessment of explanation quality. Ultimately, developing unified metrics that integrate coherence, correctness, and relevance would result in a more robust proxy for human-centered evaluation standards. More broadly, we envision applying this framework to a wide range of domains beyond climate misinformation. Scientific question answering, medical diagnostics, and legal justification all require explanations that are both causally sound and resource-efficient, making them well-suited for the proposed approach. 

\newpage
\bibliography{main}{}

\bibliographystyle{plain}

\end{document}